\documentclass[runningheads]{llncs}

\usepackage{eccv}

\usepackage{eccvabbrv}

\usepackage{graphicx}
\usepackage{booktabs}
\usepackage{multirow}
\usepackage{algorithm}
\usepackage{algpseudocode}
\usepackage[table]{xcolor}
\colorlet{rebuttalcolor}{black} % Change black to blue to restore rebuttal highlights.
\usepackage{float}

\usepackage[accsupp]{axessibility}  % Improves PDF readability for those with disabilities.

\usepackage{hyperref}

\usepackage{orcidlink}

\begin{document}

% ---------------------------------------------------------------
\title{Beyond Random Sampling: Distribution-Aware Alignment for Semi-Supervised Medical Image Segmentation} 

\titlerunning{Distribution-Aware Alignment for SSMIS}

\author{Weihao Yan\inst{1}\orcidlink{0000-0002-7838-0790} \and
Yeqiang Qian\inst{1}\thanks{Corresponding author.}\orcidlink{0000-0003-0831-8702} \and
Yi Dong\inst{2}\orcidlink{0000-0002-0212-1477} \and
Ming Yang\inst{1}\orcidlink{0000-0002-8679-9137}}

\authorrunning{W.~Yan et al.}

\institute{School of Automation and Intelligent Sensing, Shanghai Jiao Tong University, Shanghai, China \\
\email{qianyeqiang@sjtu.edu.cn}
\and
Department of Ultrasound, Xinhua Hospital Affiliated to Shanghai Jiao Tong University School of Medicine, Shanghai, China
}

\maketitle

\begin{abstract}
Precise medical image segmentation is crucial for clinical diagnosis and treatment planning, yet relies heavily on expensive expert annotations. Semi-supervised medical image segmentation (SSMIS) offers a cost-effective solution but typically operates under the assumption of independent and identically distributed (i.i.d.) data, defaulting to random sampling. While statistically valid at scale, this strategy suffers from severe representation bias in low-data regimes, failing to capture the heterogeneous medical data manifold. To address this, we propose a highly data-efficient framework driven by distribution alignment. First, we introduce an offline Distribution-Aware Sample Selection strategy. By leveraging Vision Foundation Models (VFMs) and our designed Density-K-Center algorithm, we explicitly identify representative structural anchors, establishing a more representative labeled domain. Second, to bridge the remaining distribution gap, we propose the Memory-guided Copy-Paste (MCP) module. Tailored for the inherent class imbalance in medical scans, MCP leverages a semantic memory mechanism to retrieve historically consistent priors for cross-domain alignment, encouraging semantic consistency. Coupled with an easy-to-hard progressive schedule, this framework effectively mitigates early-stage pseudo-label noise. Extensive experiments on six diverse 2D and 3D datasets demonstrate strong segmentation performance, particularly in extremely low-labeled scenarios (\eg, 1/16 ratio).
\end{abstract}

% \vspace{-1em}
\section{Introduction}
\label{sec:intro}
% \vspace{-1em}
Medical image segmentation is a fundamental task for modern healthcare, enabling precise quantification of anatomical structures for diagnosis and therapy planning \cite{abdominal_seg,cross}. However, training deep segmentation models requires massive pixel-level annotations, necessitating expensive and time-consuming expert labor. Semi-supervised medical image segmentation (SSMIS) has emerged as a pivotal solution, aiming to leverage limited labeled data alongside abundant unlabeled scans using various strategies such as consistency regularization, self-training, and co-training\cite{BCP,ABD,midss,synfoc,unimatch,unimatchv2,cross_pseudo}.

\begin{figure}
  \centering
  \includegraphics[width=1.0\textwidth]{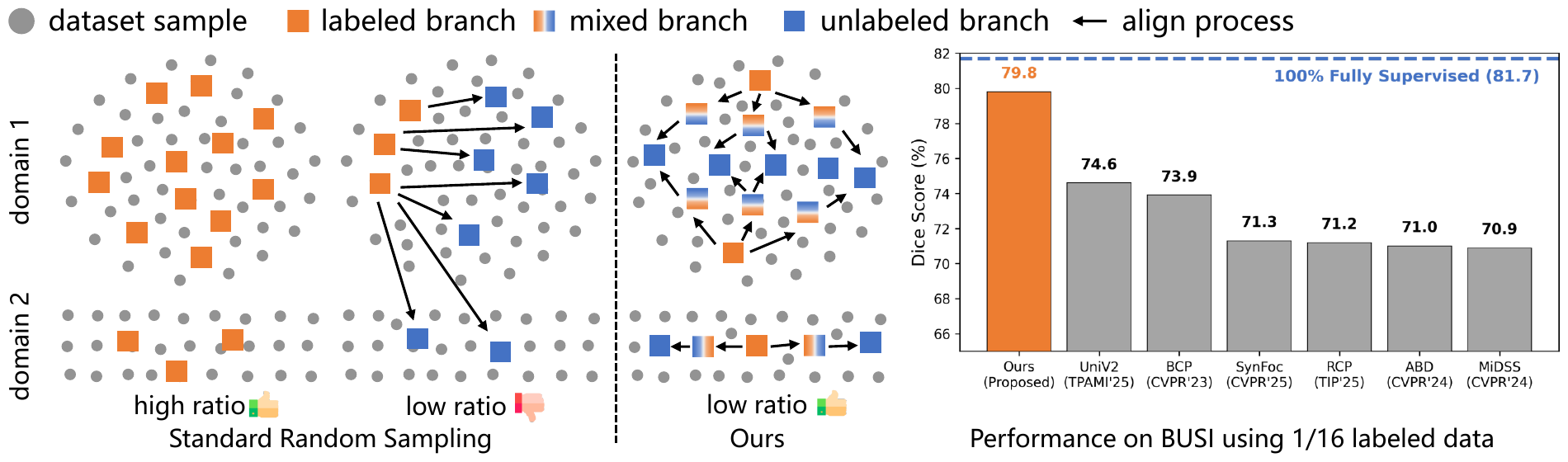}
  \caption{{Motivation and Performance Overview.} {Left:} Standard random sampling covers the manifold at a high labeled ratio but suffers from severe representation bias at a low ratio. {Middle:} Our method anchors globally representative samples and aligns domains via a mixed branch in progressive schedule.
  {Right:} Dice Score comparison on the BUSI dataset (1/16 ratio). Our approach outperforms recent SOTA methods and narrows the gap to the fully supervised setting.}
  \label{fig:motivation}
\end{figure}

Despite notable advances, most existing SSMIS frameworks assume labeled and unlabeled data are independent and identically distributed (i.i.d.) \cite{unimatch,unimatchv2,semivl}, relying on random sampling for initialization. As illustrated in \cref{fig:motivation}, while random sampling provides adequate coverage with ample annotations (high ratio), it exhibits severe instability in {low-data regimes}. In constrained scenarios, randomly selected samples clump into specific sub-clusters, failing to cover the global data manifold and causing representation bias. Conversely, our method anchors broader regions of the manifold and explicitly aligns features across domains. This enables our framework to outperform standard methods \cite{unimatchv2,BCP,ABD,synfoc,RCP,midss}, approaching fully supervised performance (\cref{fig:motivation}, right).

\vspace{-1.5em}
\vspace{-1.5em}

Drawing inspiration from domain alignment principles\cite{da_survey,da_analysis,meduda_survey,wang2019semi}, we argue that effective SSMIS in low-data regimes requires explicitly addressing the distribution shift between the sparse labeled set (source) and the diverse unlabeled set (target). Rather than treating them as a single uniform domain, our objective is twofold: first, to establish a representative source domain that anchors the global manifold, and second, to robustly bridge the cross-domain gap during the learning phase. 
To this end, we propose a highly data-efficient framework driven by two core innovations. First, we introduce an offline strategy named \textbf{Distribution-Aware Sample Selection}. Instead of blind selection, we leverage the semantic priors of Vision Foundation Models (VFMs) such as DINOv2 \cite{dinov2,dinov3} to map clinical scans into a high-dimensional feature manifold. We then employ the designed {Density-K-Center} algorithm to select structural anchors. By systematically balancing global diversity with local typicality, this strategy ensures the initial labeled set serves as a less biased and reliable source domain, even with minimal annotations.

Second, we propose the \textbf{Memory-guided Copy-Paste (MCP)} module to facilitate robust feature alignment. Conventional mixing strategies like CutMix\cite{cutmix,BCP,ABD,midss,RCP} often fail on medical data due to extreme class imbalance—small anatomical targets (foreground) are easily overwhelmed by large background regions. MCP addresses this by maintaining {category-specific Semantic Memory Banks} (explicitly separating foreground and background queues). This design allows us to actively retrieve and paste historically consistent foreground patches onto unlabeled images, enhancing the model's sensitivity to small anatomical structures while ensuring semantic consistency. Coupled with an easy-to-hard progressive activation schedule, MCP effectively prevents the model from overfitting to noisy pseudo-labels in early training stages.

The main contributions of this work are summarized as follows:
\begin{itemize}
  \item We propose the {Distribution-Aware Sample Selection} strategy to mitigate the representation bias of random sampling in low-data regimes. By leveraging VFM embeddings and Density-K-Center clustering, it identifies representative structural anchors to establish a less biased labeled domain.  % for both 2D and 3D data.
  \item We introduce the {Memory-guided Copy-Paste (MCP)} module to facilitate robust cross-domain alignment. It employs category-specific memory banks to address class imbalance and integrates an easy-to-hard progressive schedule to suppress early-stage pseudo-label noise.
  \item Extensive experiments across six diverse 2D and 3D medical datasets (Ultrasound, X-ray, Dermoscopy, and MRI) demonstrate that our method consistently sets new state-of-the-art. Notably, our sampling module functions as a plug-and-play performance multiplier for existing SSMIS methods.
\end{itemize}

\section{Related Work}
% \vspace{-1em}
\subsection{Semi-Supervised Medical Image Segmentation}
SSMIS reduces reliance on expert annotations by combining limited labeled data with abundant unlabeled scans. Existing methods largely follow three paradigms: (i) \emph{consistency regularization}~\cite{fixmatch,unimatch,unimatchv2,ABD} enforce invariance; (ii) \emph{self-training}~\cite{st++,beyond,inconsistency_aware} refine pseudo-labels; and (iii) \emph{co-training}~\cite{ucmt,synfoc,cross_training,cross_pseudo} exploit consensus. However, most frameworks assume i.i.d.\ distributions, ignoring shifts from heterogeneous clinical acquisition~\cite{da_survey,da_analysis}. Under such heterogeneity, standard random sampling~\cite{midss,synfoc,bai2017semi,ssada} causes severe representation bias in {low-data regimes}. While valid at scale, it fails to anchor the global manifold with scarce labels, weakening initial supervision and degrading performance.

\vspace{-1em}
\subsection{Domain Adaptation in Medical Imaging}
UDA addresses distribution mismatches between labeled source and unlabeled target domains. Representative medical UDA techniques include image-level transfer (\eg, style translation)~\cite{cyclegan,fda,tufl}, adversarial feature alignment~\cite{advent,adapt_structured,adversarial_medical}, and self-training-based adaptation~\cite{self_training_medical,sam4udass,zhang2020collaborative,zheng2021rectifying}. While UDA explicitly models domain gaps, standard SSMIS pipelines rarely treat the labeled/unlabeled split as a domain shift problem, despite gaps naturally emerging from heterogeneous acquisition\cite{midss}. This mismatch between data reality and the i.i.d.\ assumption motivates revisiting SSMIS from a UDA-inspired perspective.

\vspace{-1em}
\subsection{Copy-Paste for Medical Image Segmentation}
Copy-Paste (CP) is a standard augmentation in SSMIS. Intuitively, it cuts areas from a source image and pastes them onto a target to synthesize diverse training samples. BCP~\cite{BCP} pioneered bi-directional CP, while later works added semantic constraints~\cite{ABD,RCP} or curriculum learning~\cite{AdaMix}. However, existing approaches face two medical-specific challenges. First, networks are optimized using small image mini-batches. Due to extreme class imbalance, within-batch mixing often fails to retrieve sparse foregrounds---as a single batch might contain no lesions---leading to severe background dominance. Second, aggressive mixing with noisy pseudo-labels early in training causes overfitting to incorrect supervision. %, exacerbating confirmation bias.
\vspace{-0.8em}

\section{Methodology}
\begin{figure}[htbp]
  \vspace{-1em}
  \centering
  \includegraphics[width=\textwidth]{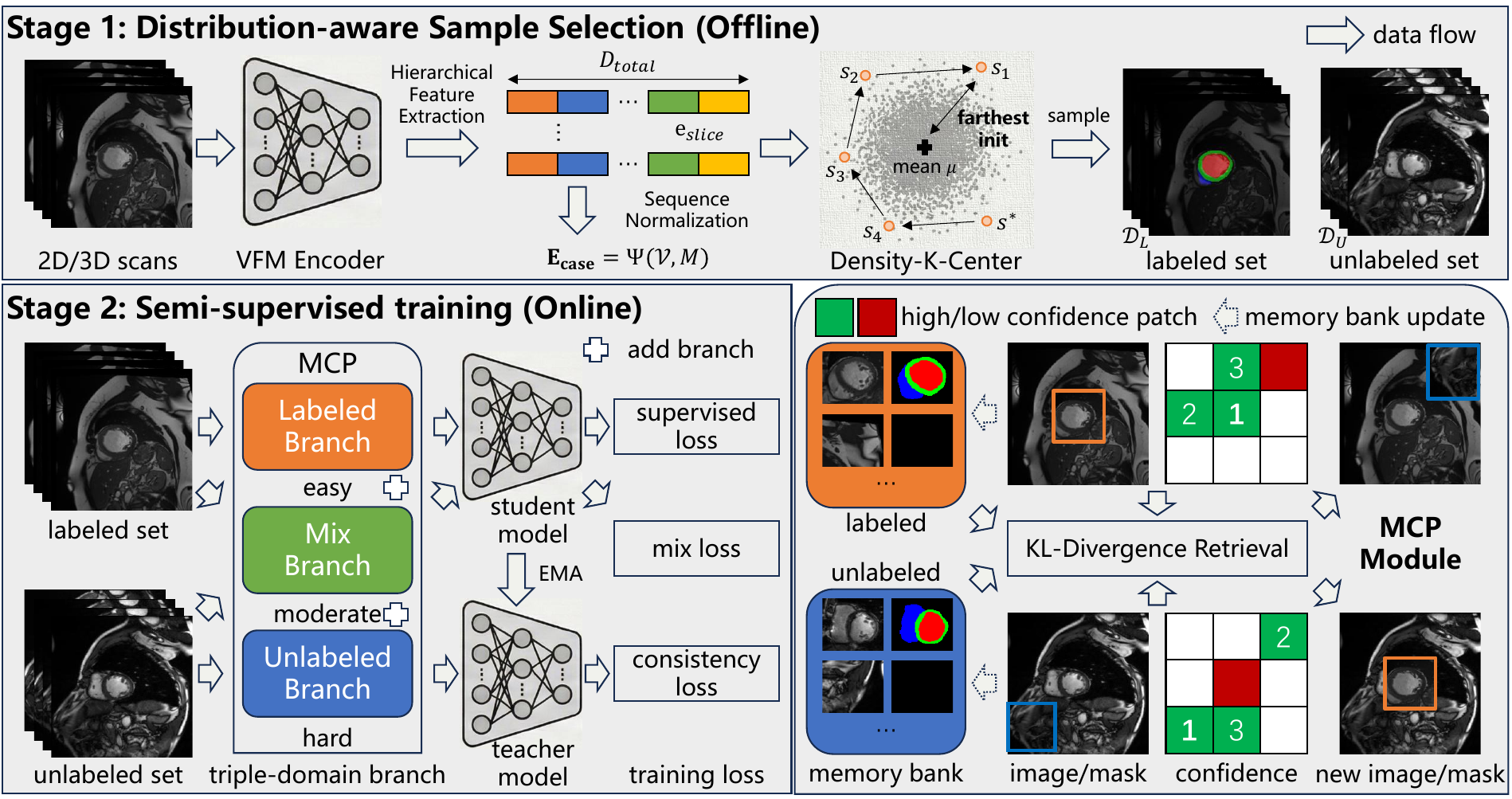}
  \caption{Overview of our framework. \textbf{Top (Stage 1):} Offline Distribution-Aware Sample Selection. Hierarchical VFM features and Density-K-Center clustering identify representative anchors covering the global manifold. \textbf{Bottom Left (Stage 2):} Online training pipeline. A teacher-student architecture optimizes labeled, mixed, and unlabeled branches. \textbf{Bottom Right:} The MCP module. A category-specific Semantic Memory Bank retrieves historical patches via KL-Divergence for cross-domain alignment.}
  \label{fig:framework}
\end{figure}

As illustrated in \cref{fig:framework}, we propose a data-efficient SSMIS framework that operates in two stages to address representation bias and mitigate pseudo-label noise. 
\textbf{Stage 1 (Offline)} establishes a globally representative labeled set via \textbf{Distribution-Aware Sample Selection}, utilizing frozen VFM priors to anchor the data manifold. 
\textbf{Stage 2 (Online)} aligns cross-domain features through the \textbf{Memory-guided Copy-Paste (MCP)} module. By integrating a semantic memory bank with an easy-to-hard progressive schedule, this stage robustly bridges the inherent distribution gap.

\vspace{-1em}
\subsection{Preliminaries and Notations}
Let the dataset be $\mathcal{D} = \mathcal{D}_L \cup \mathcal{D}_U$, where $\mathcal{D}_L = \{(x_i, y_i)\}_{i=1}^{N_L}$ represents the labeled set and $\mathcal{D}_U = \{x_j\}_{j=1}^{N_U}$ the unlabeled set. SSMIS aims to learn a robust segmentation model $f_\theta$ that accurately predicts dense masks across all data. Our UDA-inspired framework explicitly formulates $\mathcal{D}_L$ as the ``source domain'' and $\mathcal{D}_U$ as the ``target domain.'' We aim to bridge their inherent distribution gap through strategic sample selection prior to training, followed by effective cross-domain alignment during the semi-supervised learning phase.

\vspace{-1em}
\subsection{Distribution-Aware Sample Selection}
% To address the limitations and potential bias of random sampling in SSMIS, we propose a representative sample selection strategy. 
Using a pre-trained Vision Foundation Model (VFM), we project the raw clinical data into a high-dimensional feature space $\mathcal{M}$, where anatomical similarities can be easily measured by feature distances.

\vspace{-1em}
\subsubsection{Hierarchical Feature Encoding}
For a 2D medical image $x \in \mathbb{R}^{H \times W \times 3}$, we extract multi-level semantic features using a frozen VFM encoder $f_{vfm}$. To capture both fine-grained local textures and broader global anatomical context, we select a subset of $L$ intermediate layers $\mathcal{L} = \{l_1, l_2, \dots, l_L\}$. 

Let $\mathbf{T}_l \in \mathbb{R}^{N_p \times D_l}$ be the patch tokens extracted from layer $l \in \mathcal{L}$, where $N_p$ is the number of spatial patches and $D_l$ is the feature dimension. The encoding process aggregates these patch tokens spatially, concatenates them channel-wise, and applies $L_2$ normalization to obtain the final slice embedding $\mathbf{e}_{slice}$:
\begin{equation}
  \mathbf{z}_l = \text{GAP}(\mathbf{T}_l), \quad \mathbf{e}_{raw} = \text{Concat}(\mathbf{z}_{l_1}, \dots, \mathbf{z}_{l_L}), \quad
  \mathbf{e}_{slice} = \frac{\mathbf{e}_{raw}}{\|\mathbf{e}_{raw}\|_2}
\end{equation}
where $\text{GAP}(\cdot)$ denotes Global Average Pooling, and $\mathbf{e}_{raw} \in \mathbb{R}^{D_{total}}$ with $D_{total} = \sum_{i=1}^L D_{l_i}$. This normalization projects all embeddings onto a unit hypersphere, making the feature space suitable for cosine-distance-based similarity comparison. For instance, using DINOv2-Small with $L=4$ layers and $D_l=384$ yields a unified descriptor $\mathbf{e}_{slice} \in \mathbb{R}^{1536}$.

\vspace{-1em}
\subsubsection{Volumetric Sequence Normalization}
In clinical practice, 3D medical modalities (\eg, MRI and CT) provide volumetric data consisting of multiple sequential 2D slices. To treat each patient case as a coherent entity, we propose a Case-Sequence Feature Normalization strategy. 

Let a variable-length case be defined as $\mathcal{V} = \{\mathbf{e}_n\}_{n=1}^N$. Directly aggregating these slices (\eg, via mean pooling) risks mixing critical pathological details into a diluted single vector. Instead, we map each variable-length sequence into a standardized feature tensor $\mathbf{E}_{case} \in \mathbb{R}^{M \times D_{total}}$, where $M$ is a fixed reference sequence length:
\begin{equation}
  \mathbf{E}_{case} = \Psi \left( \mathcal{V}, M \right)
\end{equation}
% \vspace{l}
To implement the normalization function $\Psi$, we exclusively employ a {Structural Centering} strategy\cite{center}. Specifically, we extract the $M$ central slices of the volume. This simple yet effective design aligns with clinical practice: it focuses on the core anatomical regions where lesions or target organs are most frequently located, while ignoring uninformative background slices at the ends. The resulting tensor $\mathbf{E}_{case}$ successfully preserves the patient's structural information, allowing our selection algorithm to evaluate the scan as a whole.

\subsubsection{Density-Weighted K-Center Sampling}
To construct the labeled set $\mathcal{D}_L$, we introduce the Density-K-Center algorithm to balance feature diversity and typicality. We first estimate a local density weight $w_i$ for each sample $i$ based on its $k$-nearest neighbors ($\text{KNN}$). In our implementation, we set $k = 20$. The local density is computed as:
\begin{equation}
  \rho_i = \frac{1}{k} \sum_{j \in \text{KNN}(i)} \left\| \mathbf{e}_i - \mathbf{e}_j \right\|_2^2, \quad w_i = \sigma \left( \frac{1}{\rho_i + \epsilon} \right)
\end{equation}
where $\left\| \mathbf{e}_i - \mathbf{e}_j \right\|_2^2$ is the squared Euclidean distance, $\epsilon$ is a small constant (\eg, $10^{-6}$) to prevent zero division, and $\sigma(\cdot)$ scales the weights into a stable range $[w_{\text{min}}, 1.0]$. We empirically set $w_{\text{min}} = 0.3$ to ensure the weight acts as a soft penalty rather than a hard threshold, preventing extreme bias toward uninformative outlier samples.

\paragraph{Initialization via Farthest-from-Mean}
To start the selection process, the first sample is chosen as the one with the maximum distance from the global geometric centroid of the dataset:
\begin{equation}
  s_1 = \arg \max_{i} \left( \text{dist}(\mathbf{e}_i, \mu) \right), \quad \mu = \frac{1}{N} \sum_{i=1}^{N} \mathbf{e}_i
\end{equation}
where $\mu$ is the mean feature vector of the entire dataset. This ensures the sampling begins from a distinct boundary of the data distribution. 

\paragraph{Greedy Selection Process}
At each subsequent step, we select the next sample $s^*$ that maximizes the density-weighted distance to the previously selected samples:
\begin{equation}
  s^* = \arg \max_{i \notin \mathcal{D}_L} \left( w_i \cdot \min_{s \in \mathcal{D}_L} \text{dist}(\mathbf{e}_i, \mathbf{e}_s) \right) 
\end{equation}
where $\mathbf{e}_i$ and $\mathbf{e}_s$ denote the feature embeddings of a candidate sample from the unselected pool and an already selected anchor, respectively. This iterative process continues until the annotation budget $N_L$ is met. The algorithm encourages samples from diverse regions to be selected, while the density weight $w_i$ guides the model to prefer denser regions where the clinical data is most concentrated.

\subsection{Memory-guided Copy-Paste with Progressive Alignment}
To bridge the distribution gap between labeled and unlabeled domains, we propose the Memory-guided Copy-Paste (MCP) module. MCP aligns features by combining spatial augmentation with a foreground-aware Memory Bank and a progressive training schedule.

\subsubsection{Triple-Domain Branch Design}
Unlike existing methods focusing on single-domain or bidirectional augmentation \cite{BCP, ABD, RCP}, our MCP operates across three branches tailored for medical images:
\begin{itemize}
    \item \textbf{Labeled Domain Branch}: Performs augmentation strictly within $\mathcal{D}_L$ to establish a strong supervised baseline.
    \item \textbf{Unlabeled Domain Branch}: Conducts augmentation within $\mathcal{D}_U$ using pseudo-labels that generated from the teacher model, encouraging learning from ambiguous tissue regions.
    \item \textbf{Mixed Domain Branch}: Executes cross-domain Copy-Paste ($\mathcal{D}_L \leftrightarrow \mathcal{D}_U$), transferring reliable anatomical supervision between domains.
\end{itemize}

An image $x \in \mathbb{R}^{H \times W \times 3}$ is uniformly divided into a $7 \times 7$ patch grid. For each active branch, the loss combines Dice and Cross-Entropy (CE) losses:
\begin{equation}
  \mathcal{L}_{branch} = \frac{1}{2} \left( \mathcal{L}_{dice}(\hat{y}, y, \omega) + \mathcal{L}_{ce}(\hat{y}, y, \omega) \right)
\end{equation}
where $\hat{y}$ is the prediction and $y$ the ground truth ($\mathcal{D}_L$) or pseudo-label ($\mathcal{D}_U$). The confidence weight $\omega$ is 1.0 for ground truth regions and 0.5 for pseudo-labeled regions to mitigate noise propagation.

\subsubsection{Semantic Memory Bank with Foreground-Background Separation}
The medical images exhibit severe class imbalance, with sparse regions of interest against expansive backgrounds. To overcome this and limited intra-batch diversity, we decouple our Patch Memory Bank into foreground and background subsets: $\mathcal{B} = \{\mathcal{B}^{fg}, \mathcal{B}^{bg}\}$. Each entry $(\mathbf{p}, \mathbf{y}, \mathbf{d}, c, id)$ stores the patch pixels, mask, class probability, mean confidence, and source ID. Target-class patches route to $\mathcal{B}^{fg}$; healthy context patches route to $\mathcal{B}^{bg}$.

\paragraph{Hybrid Eviction Strategy} 
When subsets reach memory bank capacity $C_{bank}$ (1024), we independently apply a hybrid eviction policy: 50\% First-In-First-Out (FIFO) for freshness, and 50\% removing entries with the lowest confidence $c$. This class-separated updating prevents majority background patches from flushing out rare foreground anomalies.

\paragraph{Semantic Retrieval} 
During augmentation, MCP selectively retrieves patches by foreground/background identity with probability $P_{bank}$, while the rest are randomly sampled from the current mini-batch. 
To synthesize anomalies, a foreground lesion from $\mathcal{B}^{fg}$ can replace a background region. Given a target patch class distribution $\mathbf{d}_t$, we retrieve the most semantically similar source patch $\mathbf{p}_s^*$ by minimizing the Kullback-Leibler (KL) divergence:
\begin{equation}
  \mathbf{p}_s^* = \arg \min_{\substack{\mathbf{p} \in \mathcal{B}, id \neq id_t}} D_{KL}(\mathbf{d}_s \| \mathbf{d}_t), \quad D_{KL}(\mathbf{d}_s \| \mathbf{d}_t) = \sum_i d_{s,i} \log \left( \frac{d_{s,i}}{d_{t,i}} \right)
\end{equation}
This foreground-aware matching ensures retrieved patches naturally fit the target's tissue context, preventing unnatural semantic mixing.

\subsubsection{Progressive Branch Activation}
To ensure training stability, we use an easy-to-hard progressive activation schedule over a ramp-up period $E_{prog} = R_{prog} \times E_{total}$, where $R_{prog}$ is the epoch ratio and $E_{total}$ is the total number of epochs:
\begin{itemize}
    \item \textbf{Phase 1} ($0 \le e < \frac{1}{2} E_{prog}$): Only the Labeled Domain Branch is active for a stable baseline.
    \item \textbf{Phase 2} ($\frac{1}{2} E_{prog} \le e < E_{prog}$): The Mixed Domain Branch is introduced for cross-domain alignment.
    \item \textbf{Phase 3} ($e \ge E_{prog}$): All branches are fully activated.
\end{itemize}
The overall training objective at epoch $e$ is:
\begin{equation}
  \mathcal{L}_{MCP} = \mathcal{L}_{labeled} + \mathbb{I}\left(e \ge \frac{1}{2} E_{prog}\right) \mathcal{L}_{mixed} + \mathbb{I}(e \ge E_{prog}) \mathcal{L}_{unlabeled}
\end{equation}
where $\mathbb{I}(\cdot)$ is an indicator function. This prevents early overfitting to noisy pseudo-labels. The algorithm process is detailed in supplementary material.

\section{Experiments}
\label{sec:experiments}

\subsection{Datasets and Evaluation Metrics}
We evaluate our framework on six public diverse datasets.
\paragraph{2D Datasets.}
(1) \textbf{BUSI \cite{busi}}: 780 breast ultrasound images (623/157 split) with benign, malignant, and normal cases.
(2) \textbf{PMTCXR \cite{pmtcxr}}: 2,669 chest X-ray images (2,379/290 split) for pneumothorax segmentation.
(3) \textbf{ISIC \cite{isic}}: 2,594 dermoscopy images (1,815/779 split) containing complex skin lesions.
(4) \textbf{CAMUS \cite{camus}}: Cardiac ultrasound sequences from 500 patients (2,800/800 images) annotated for LV and LA structures.

\paragraph{3D Datasets.}
(1) \textbf{PROMISE \cite{promise12}}: Prostate MRI volumes from 50 patients (split 35/5/10).
(2) \textbf{ACDC \cite{ACDC}}: Cardiac cine-MRI from 100 patients (70/10/20 split), focusing on RV, LV, and Myocardium segmentation.
For both 3D datasets, volumes are converted into 2D slices for training convenience \cite{bai2017semi}, but case-level evaluation is used by concatenating the slice predictions back to 3D volumes.

\paragraph{Evaluation Metrics}
We employ the Dice Similarity Coefficient (DSC) and Intersection over Union (IoU) to measure regional overlap (higher is better), alongside the 95\% Hausdorff Distance (HD95) and Average Surface Distance (ASD) to evaluate boundary localization accuracy (lower is better).

\subsection{Implementation Details}
\label{sec:implementation}

Following UniV2\cite{unimatchv2}, we adopt the DPT decoder \cite{dpt} with a DINOv2-Small backbone \cite{dinov2}. Training inputs are resized to $518 \times 518$. Model is optimized using AdamW (weight decay $0.01$) with poly-decayed learning rates ($5 \times 10^{-6}$ for encoder, $2 \times 10^{-4}$ for decoder). Models are trained on two RTX 4090 GPUs with balanced labeled/unlabeled mini-batches, using OHEM, cross-entropy, and Dice losses.
For offline selection, we apply Density-K-Center ($k=20$) on $L_2$-normalized tokens from four intermediate DINOv2 layers. The MCP module employs a $7 \times 7$ grid, memory capacity $C_{bank} = 1024$, and retrieval probability $P_{bank} = 0.2$, matching optimal patches via KL-divergence. After a 10\% epoch ramp-up ($R_{prog} = 0.1$), all branches share equal loss weights. For fair comparisons, predictions are resized to $256 \times 256$ (ACDC) or $224 \times 224$ (others) before metric calculation. Full source code will be available at \url{https://github.com/ywher/DAA4SSMIS}.

% \vspace{-2em}
\subsection{Comparison with State-of-the-Art Methods}

\subsubsection{Results on 3D Benchmarks (PROMISE, ACDC)}
\cref{tab:sota_3d} details the results on 3D volumetric data.
On \textbf{PROMISE}, where random sampling causes severe bias at the 1/16 ratio (only 2 cases), our Distribution-Aware Selection identifies representative volumetric anchors, achieving 87.3\% DSC and outperforming UniV2 by 5.3\%.
On \textbf{ACDC}, despite training on 2D slices, our method preserves strong volumetric consistency. At the 1/20 ratio (3 cases), we surpass the Labeled-Only baseline by 8.1\% DSC. Even as performance saturates at the 1/5 ratio, our method further improves performance, reaching 91.5\% DSC with low boundary error (ASD 0.3).
\vspace{-1.5em}
\begin{table}[H]
\centering
\caption{Comparison with SOTA methods on \textbf{3D Datasets} (PROMISE, ACDC). Specific labeled ratios (Cases) are defined in the section headers. Labeled Only: the supervised baseline trained solely on the labeled set.
Best: \textbf{Bold}, Second: \underline{Underline}.}
\label{tab:sota_3d}
\setlength{\tabcolsep}{3pt}
\resizebox{\textwidth}{!}{
\begin{tabular}{l c c cc cc cc}
\toprule
\multirow{3}{*}{Methods} & \multirow{3}{*}{Venue} & \multirow{3}{*}{Encoder} & \multicolumn{6}{c}{Ratio and the number of labeled cases} \\
\cmidrule(lr){4-9}
& & & \multicolumn{2}{c}{Low Ratio} & \multicolumn{2}{c}{Medium Ratio} & \multicolumn{2}{c}{High Ratio} \\
\cmidrule(lr){4-5} \cmidrule(lr){6-7} \cmidrule(lr){8-9}
& & & IoU/DSC$\uparrow$ & ASD/HD$\downarrow$ & IoU/DSC$\uparrow$ & ASD/HD$\downarrow$ & IoU/DSC$\uparrow$ & ASD/HD$\downarrow$ \\
\midrule
% ================= PROMISE =================
\multicolumn{9}{c}{\cellcolor{gray!10}\textbf{Dataset: PROMISE} (Ratios: 1/16, 1/10, 1/5 $\mid$ Cases: 2, 3, 7)} \\
\midrule
BCP\cite{BCP} & CVPR'23 & UNet & 60.5/73.1 & 1.2/4.6 & 61.2/74.9 & 3.0/11.8 & 65.6/78.7 & 7.0/20.6 \\
MiDSS\cite{midss} & CVPR'24 & UNet & 66.2/79.4 & 1.1/3.4 & 71.6/83.3 & \underline{1.0}/2.7 & 70.5/82.5 & 1.1/2.8 \\
ABD\cite{ABD} & CVPR'24 & SwinUNet & 65.7/79.1 & 1.4/3.6 & 68.4/80.9 & 1.3/3.4 & 68.0/80.7 & 1.4/3.5 \\
RCP\cite{RCP} & TIP'25 & UNet & 68.0/80.9 & 2.9/9.3 & 70.2/82.4 & 1.7/5.1 & 68.2/80.8 & 2.9/8.9 \\
AdaMix\cite{AdaMix} & MedIA'25 & UNet & 55.1/70.2 & 1.7/5.1 & 66.0/79.4 & 2.4/4.9 & 61.8/76.2 & 2.9/8.7 \\
SynFoc\cite{synfoc} & CVPR'25 & MedSAM & \underline{75.3}/\underline{85.7} & \underline{0.9}/\underline{2.8} & 75.9/86.2 & 1.1/3.6 & 74.4/85.3 & \underline{0.9}/2.7 \\
UniV2\cite{unimatchv2} & TPAMI'25 & DINOv2-S & 69.8/82.0 & \underline{0.9}/3.2 & \underline{76.7}/\underline{86.7} & \textbf{0.7}/\underline{2.4} & \underline{75.6}/\underline{86.0} & \textbf{0.8}/\underline{2.6} \\ \midrule
Labeled Only & - & DINOv2-S & 64.0/77.8 & 1.3/4.3 & 67.3/79.8 & 1.1/3.5 & 67.5/80.2 & 1.1/3.4 \\
\textbf{Ours} & - & DINOv2-S & \textbf{77.7/87.3} & \textbf{0.7/2.3} & \textbf{78.4/87.9} & \textbf{0.7/2.3} & \textbf{77.5/87.2} & \textbf{0.8/2.3} \\
\midrule
% ================= ACDC =================
\multicolumn{9}{c}{\cellcolor{gray!10}\textbf{Dataset: ACDC} (Ratios: 1/20, 1/10, 1/5 $\mid$ Cases: 3, 7, 14)} \\
\midrule
BCP\cite{BCP} & CVPR'23 & UNet & 75.6/85.6 & 1.6/5.3 & 80.4/88.7 & 0.7/2.4 & 83.0/90.4 & 0.5/1.8 \\
MiDSS\cite{midss} & CVPR'24 & UNet & \underline{80.6}/\underline{88.9} & 0.8/3.5 & 82.0/89.8 & \underline{0.5}/\underline{1.5} & \underline{83.6}/90.8 & \textbf{0.3}/\underline{1.2} \\
ABD\cite{ABD} & CVPR'24 & SwinUNet & 78.4/87.4 & 0.6/1.7 & 82.2/90.0 & \underline{0.5}/\textbf{1.3} & 83.0/90.5 & 0.7/2.7 \\
RCP\cite{RCP} & TIP'25 & UNet & 79.0/87.9 & 1.4/5.1 & 81.5/89.5 & 1.3/6.2 & 82.7/90.3 & 1.0/3.0 \\
AdaMix\cite{AdaMix} & MedIA'25 & UNet & 77.2/86.7 & 1.2/4.7 & 78.0/87.2 & 1.5/5.2 & 81.1/89.2 & 0.9/2.0 \\
SynFoc\cite{synfoc} & CVPR'25 & MedSAM & 73.9/84.3 & 2.1/6.8 & 77.9/87.2 & 0.9/2.5 & 79.5/88.3 & 0.8/3.1 \\
UniV2\cite{unimatchv2} & TPAMI'25 & DINOv2-S & 80.2/88.6 & \underline{0.5}/\underline{1.6} & \underline{82.9}/\underline{90.4} & \underline{0.5}/1.6 & \underline{83.6}/\underline{90.9} & \underline{0.4}/1.5 \\ \midrule
Labeled Only & - & DINOv2-S & 70.9/82.3 & 4.1/15.4 & 79.3/88.1 & 1.5/5.3 & 81.9/89.8 & 0.6/2.6 \\
\textbf{Ours} & - & DINOv2-S & \textbf{82.8/90.4} & \textbf{0.3/1.2} & \textbf{83.8/91.0} & \textbf{0.4/1.3} & \textbf{84.6/91.5} & \textbf{0.3/1.1} \\
\bottomrule
\end{tabular}
}
\end{table}
% \vspace{-1.5em}

\subsubsection{Results on 2D Benchmarks (BUSI, PMTCXR, ISIC, CAMUS)}
\cref{tab:sota_2d} summarizes performance across four datasets.
On \textbf{BUSI}, our framework excels at the scarce 1/16 ratio, achieving 79.8\% DSC and reducing HD95 by 12.6 pixels vs. UniV2.
On the heterogeneous \textbf{PMTCXR}, many SSMIS methods collapse ($<40\%$ DSC) at the 1/16 ratio, whereas our method maintains 52.9\% DSC.
For \textbf{ISIC}, our method outperforms UniV2 by 1.9\% DSC at the 1/40 ratio.
Finally, on \textbf{CAMUS}, we reach 93.2\% DSC (1/4 ratio), supporting our method's efficacy on structure-consistent cardiac ultrasounds.

\vspace{-2em}
\begin{table}[H]
\centering
\caption{Comparison with SOTA methods on \textbf{2D Datasets} (BUSI, PMTCXR, ISIC, CAMUS). The specific labeled ratios and counts are indicated in the section headers. }  % for each dataset 
% \vspace{-1em}
\label{tab:sota_2d}
\setlength{\tabcolsep}{3pt}
\resizebox{\textwidth}{!}{
\begin{tabular}{l c c cc cc cc}
\toprule
\multirow{3}{*}{Methods} & \multirow{3}{*}{Venue} & \multirow{3}{*}{Encoder} & \multicolumn{6}{c}{Ratio and the number of labeled images} \\
\cmidrule(lr){4-9}
& & & \multicolumn{2}{c}{Low Ratio} & \multicolumn{2}{c}{Medium Ratio} & \multicolumn{2}{c}{High Ratio} \\
\cmidrule(lr){4-5} \cmidrule(lr){6-7} \cmidrule(lr){8-9}
& & & IoU/DSC$\uparrow$ & ASD/HD$\downarrow$ & IoU/DSC$\uparrow$ & ASD/HD$\downarrow$ & IoU/DSC$\uparrow$ & ASD/HD$\downarrow$ \\
\midrule
% ================= BUSI =================
\multicolumn{9}{c}{\cellcolor{gray!10}\textbf{Dataset: BUSI} (Ratios: 1/16, 1/8, 1/4 $\mid$ Counts: 39, 78, 156)} \\
\midrule
BCP\cite{BCP} & CVPR'23 & UNet & 67.2/73.9 & \underline{33.1}/\underline{48.2} & \underline{70.9}/76.7 & 34.7/46.2 & 71.2/77.5 & 29.9/41.7 \\
MiDSS\cite{midss} & CVPR'24 & UNet & 64.3/70.9 & 42.6/56.5 & 64.2/70.0 & 39.5/54.0 & 68.2/74.8 & 32.7/45.1 \\
ABD\cite{ABD} & CVPR'24 & SwinUNet & 63.6/71.0 & 42.0/{57.6} & 66.3/73.3 & 39.5/53.5 & 69.9/76.4 & 33.1/45.0 \\
RCP\cite{RCP} & TIP'25 & UNet & 64.3/71.2 & 44.2/57.8 & 64.9/71.9 & 42.4/55.4 & 65.1/71.8 & 38.6/53.5 \\
AdaMix\cite{AdaMix} & MedIA'25 & UNet & 51.6/58.6 & 65.2/79.3 & 61.0/67.9 & 49.7/60.9 & 63.2/70.0 & 49.5/58.9 \\
SynFoc\cite{synfoc} & CVPR'25 & MedSAM & 64.8/71.3 & 39.2/54.0 & 69.0/74.9 & 35.4/47.0 & 71.8/78.1 & \underline{27.3}/37.3 \\
UniV2\cite{unimatchv2} & TPAMI'25 & DINOv2-S & \underline{67.7}/\underline{74.6} & 36.7/48.8 & 70.8/\underline{77.5} & 35.0/\underline{44.7} & 73.2/79.2 & 31.0/39.6 \\ \midrule
Labeled Only & - & DINOv2-S & 61.1/67.9 & 47.5/61.6 & 68.4/74.5 & \underline{33.8}/45.5 & \underline{73.7}/\underline{79.7} & 27.6/\underline{37.1} \\
\textbf{Ours} & - & DINOv2-S & \textbf{73.4/79.8} & \textbf{27.0/36.2} & \textbf{75.0/81.0} & \textbf{25.1/34.2} & \textbf{75.8/81.9} & \textbf{21.8/31.3} \\
\midrule
% ================= PMTCXR =================
\multicolumn{9}{c}{\cellcolor{gray!10}\textbf{Dataset: PMTCXR} (Ratios: 1/16, 1/8, 1/4 $\mid$ Counts: 149, 297, 595)} \\
\midrule
BCP\cite{BCP} & CVPR'23 & UNet & 22.8/32.8 & 43.1/81.9 & 28.0/39.1 & 28.8/65.3 & 30.9/42.5 & 24.1/59.1 \\
MiDSS\cite{midss} & CVPR'24 & UNet & 21.9/31.7 & 38.9/76.6 & 27.7/38.8 & 29.6/64.1 & 29.8/40.6 & 31.7/64.3 \\
ABD\cite{ABD} & CVPR'24 & SwinUNet & 24.8/35.9 & 29.8/69.0 & 26.8/38.2 & 26.3/63.3 & 30.4/42.2 & {19.9}/52.6 \\
RCP\cite{RCP} & TIP'25 & UNet & 23.1/33.2 & 37.4/73.8 & 24.3/34.1 & 43.8/77.3 & 30.4/41.8 & 25.7/58.5 \\
AdaMix\cite{AdaMix} & MedIA'25 & UNet & 22.4/31.9 & 24.1/64.7 & 27.1/37.8 & \underline{18.5}/57.7 & 31.4/42.9 & \underline{17.0}/53.2 \\
SynFoc\cite{synfoc} & CVPR'25 & MedSAM & 27.3/38.8 & \underline{23.1}/56.2 & 31.1/42.9 & {20.9}/53.6 & 34.1/46.3 & {19.9}/50.6 \\
UniV2\cite{unimatchv2} & TPAMI'25 & DINOv2-S & \underline{36.0}/\underline{47.6} & 26.0/\underline{52.0} & \underline{37.3}/\underline{49.0} & 25.6/52.5 & \underline{38.3}/\underline{50.3} & 21.2/\underline{45.6} \\ \midrule
Labeled Only & - & DINOv2-S & 30.0/41.5 & 23.7/56.1 & 34.7/46.8 & 21.8/\underline{50.4} & 36.6/48.6 & 20.6/47.7 \\
\textbf{Ours} & - & DINOv2-S & \textbf{40.7/52.9} & \textbf{18.9/46.0} & \textbf{40.8/53.4} & \textbf{12.8/38.3} & \textbf{41.6/54.5} & \textbf{13.8/39.5} \\
\midrule
% ================= ISIC =================
\multicolumn{9}{c}{\cellcolor{gray!10}\textbf{Dataset: ISIC} (Ratios: 1/80, 1/40, 1/20 $\mid$ Counts: 23, 45, 91)} \\
\midrule
BCP\cite{BCP} & CVPR'23 & UNet & 70.3/79.5 & 14.5/29.3 & 71.0/79.6 & 14.7/28.9 & 73.3/82.5 & 11.2/27.4 \\
MiDSS\cite{midss} & CVPR'24 & UNet & 74.2/82.7 & 10.9/26.2 & 72.6/81.3 & 11.4/27.2 & 76.0/84.5 & 8.9/21.7 \\
ABD\cite{ABD} & CVPR'24 & SwinUNet & 73.7/83.0 & 10.7/27.2 & 74.9/83.8 & 9.6/23.9 & 76.6/85.1 & 8.7/21.3 \\
RCP\cite{RCP} & TIP'25 & UNet & 74.8/83.0 & 11.4/25.0 & 74.5/83.0 & 9.1/22.5 & 76.3/84.7 & 9.1/21.4 \\
AdaMix\cite{AdaMix} & MedIA'25 & UNet & 72.7/81.8 & 9.7/22.4 & 74.8/83.4 & 8.5/20.0 & 75.9/84.0 & 7.9/19.5 \\
SynFoc\cite{synfoc} & CVPR'25 & MedSAM & 78.5/86.2 & 7.4/18.8 & 78.8/86.5 & \underline{6.5}/17.1 & 80.2/87.6 & 5.6/14.6 \\
UniV2\cite{unimatchv2} & TPAMI'25 & DINOv2-S & \underline{79.4}/\underline{87.2} & \underline{6.8}/\underline{16.7} & \underline{79.5}/\underline{87.1} & 7.0/\underline{16.2} & \underline{80.9}/\underline{88.4} & \underline{5.4}/\underline{13.1} \\ \midrule
Labeled Only & - & DINOv2-S & 75.3/83.9 & 10.3/26.8 & 76.8/85.1 & 8.3/22.1 & 79.5/87.2 & 6.2/16.1 \\
\textbf{Ours} & - & DINOv2-S & \textbf{81.5/88.7} & \textbf{5.6/14.3} & \textbf{81.7/89.0} & \textbf{5.1/13.4} & \textbf{81.8/89.0} & \textbf{5.0/12.7} \\
\midrule
% ================= CAMUS =================
\multicolumn{9}{c}{\cellcolor{gray!10}\textbf{Dataset: CAMUS} (Ratios: 1/16, 1/8, 1/4 $\mid$ Counts: 175, 350, 700)} \\
\midrule
BCP\cite{BCP} & CVPR'23 & UNet & 80.3/88.5 & 2.4/6.5 & 81.4/89.3 & 2.6/6.2 & 82.3/89.9 & 2.3/5.8 \\
MiDSS\cite{midss} & CVPR'24 & UNet & 81.7/89.5 & 2.5/5.9 & 83.2/90.5 & 1.9/5.1 & 84.3/91.2 & 1.8/4.6 \\
ABD\cite{ABD} & CVPR'24 & SwinUNet & 80.0/88.3 & 2.8/6.4 & 81.9/89.6 & 2.1/5.3 & 83.1/90.4 & 2.0/4.9 \\
RCP\cite{RCP} & TIP'25 & UNet & 81.6/89.4 & 2.2/5.8 & 83.1/90.4 & 1.9/5.2 & 83.8/90.8 & 1.9/5.0 \\
AdaMix\cite{AdaMix} & MedIA'25 & UNet & 81.2/89.2 & 2.3/5.8 & 82.0/89.8 & 2.1/5.4 & 82.9/90.3 & 2.2/5.3 \\
SynFoc\cite{synfoc} & CVPR'25 & MedSAM & 81.3/89.3 & 2.3/5.8 & 82.4/90.0 & 2.0/5.2 & 82.7/90.2 & 2.0/5.1 \\
UniV2\cite{unimatchv2} & TPAMI'25 & DINOv2-S & \underline{83.3}/\underline{90.5} & \underline{1.9}/\underline{5.0} & \underline{84.8}/\underline{91.5} & \textbf{1.7}/\underline{4.5} & \underline{86.4}/\underline{92.4} & \underline{1.5}/\underline{4.0} \\ \midrule
Labeled Only & - & DINOv2-S & 82.4/89.9 & 2.1/5.4 & 84.2/91.1 & \underline{1.8}/4.8 & 86.2/92.3 & 1.6/4.1 \\
\textbf{Ours} & - & DINOv2-S & \textbf{84.1/91.0} & \textbf{1.8/4.7} & \textbf{85.5/91.8} & \textbf{1.7/4.3} & \textbf{87.7/93.2} & \textbf{1.4/3.6} \\
\bottomrule
\end{tabular}
}
% \vspace{-1em}
\end{table}
% \vspace{-4em}

\subsection{Ablation Study of Core Components}
\label{sec:ablation}

We conduct a step-by-step additive ablation study on the BUSI dataset at the 1/16 and 1/8 labeled ratio (\cref{tab:ablation_components}).

\textbf{Effectiveness of the MCP.} 
Adding basic three-branch Copy-Paste (CP) to the UniV2 baseline moderately improves DSC (74.62\% to 75.58\%). Integrating the Semantic Memory Bank (MB) for KL-guided patch retrieval further raises DSC to 77.35\%. MB also reduces boundary errors (ASD: 32.29 to 29.13 pixels), supporting its role in preserving semantic integrity. Finally, the easy-to-hard Progressive Schedule (Prog.) mitigates early noisy pseudo-labels, improving the fully equipped MCP to 78.06\% DSC.
{\color{rebuttalcolor}
To examine why MCP works, we conduct diagnostics on the two targeted designs. MB increases valid foreground copy-paste rate from 86.4\% to 100.0\% and the retrieved foreground ratio from 60.6\% to 82.7\%, while Prog. improves early pseudo-label quality at epoch 5 by increasing confidence from 0.88 to 0.91 and reducing entropy from 0.30 to 0.25.
}

\textbf{Impact of Distribution-Aware Sample Selection (SS).} 
% Applying our SS strategy on top of MCP resolves local minima traps from random sampling. 
By selecting anchors that cover broader regions of the data manifold, SS provides a complementary improvement (\cref{tab:ablation_components}), reaching 79.80\% DSC (1/16). This supports that distribution-aware anchors benefit downstream semi-supervised learning.
{\color{rebuttalcolor}
A further PMTCXR ablation in the supplement supports the complementarity of SS and MCP: MCP improves UniV2 at 1/16 and 1/8 (36.0$\to$38.4 and 37.3$\to$39.4 IoU), and SS+MCP achieves the highest IoU (40.7/40.8).
}

\vspace{-2em}
\begin{table}[htbp]
\centering
\caption{Ablation of core components on BUSI (1/16 and 1/8 ratios). CP: basic three-branch Copy-Paste; MB: Semantic Memory Bank; Prog.: Progressive Activation Schedule; SS: Distribution-Aware Sample Selection.}
\vspace{-0.2cm}
\label{tab:ablation_components}
\setlength{\tabcolsep}{2pt} % 稍微收紧列间距
\resizebox{\textwidth}{!}{ % 自动缩放以适应页面宽度
\begin{tabular}{c c c c c | c c c c | c c c c}
\toprule
\multirow{2}{*}{Baseline} & \multirow{2}{*}{CP} & \multirow{2}{*}{MB} & \multirow{2}{*}{Prog.} & \multirow{2}{*}{SS} & \multicolumn{4}{c|}{1/16 Ratio} & \multicolumn{4}{c}{1/8 Ratio} \\
\cmidrule(lr){6-9} \cmidrule(lr){10-13}
& & & & & IoU(\%) $\uparrow$ & DSC(\%) $\uparrow$ & ASD $\downarrow$ & HD95 $\downarrow$ & IoU(\%) $\uparrow$ & DSC(\%) $\uparrow$ & ASD $\downarrow$ & HD95 $\downarrow$ \\
\midrule
\checkmark & & & & & 67.70 & 74.62 & 36.66 & 48.78 &  70.84 & 77.45 & 34.96 & 44.69 \\
\checkmark & \checkmark & & & & 69.12 & 75.58 & 32.29 & 44.17 & 71.19 & 77.35 & 26.38 & 38.02 \\
\checkmark & \checkmark & \checkmark & & & 70.96 & 77.35 & 29.13 & 40.75 & 72.65 & 78.74 & 27.29 & 37.03 \\
\checkmark & \checkmark & \checkmark & \checkmark & & 71.59 & 78.06 & 31.30 & 40.05 & 74.05 & 80.14 & 25.41 & 35.06 \\
\checkmark & & & & \checkmark & 71.45 & 78.06 & 30.55 & 40.92 & 74.12 & 79.82 & 30.08 & 38.75 \\
\checkmark & \checkmark & \checkmark & \checkmark & \checkmark & \textbf{73.40} & \textbf{79.80} & \textbf{27.00} & \textbf{36.20} & \textbf{74.99} & \textbf{81.01} & \textbf{25.10} & \textbf{34.16} \\
\bottomrule
\end{tabular}
}
\end{table}
\vspace{-3em}

\begin{figure}[H]
  \centering
  \resizebox{1.0\linewidth}{!}{
  % \vspace{-2em}
  \subfloat[Labeled ratio: 1/16]{
    \includegraphics[width=0.25\linewidth]{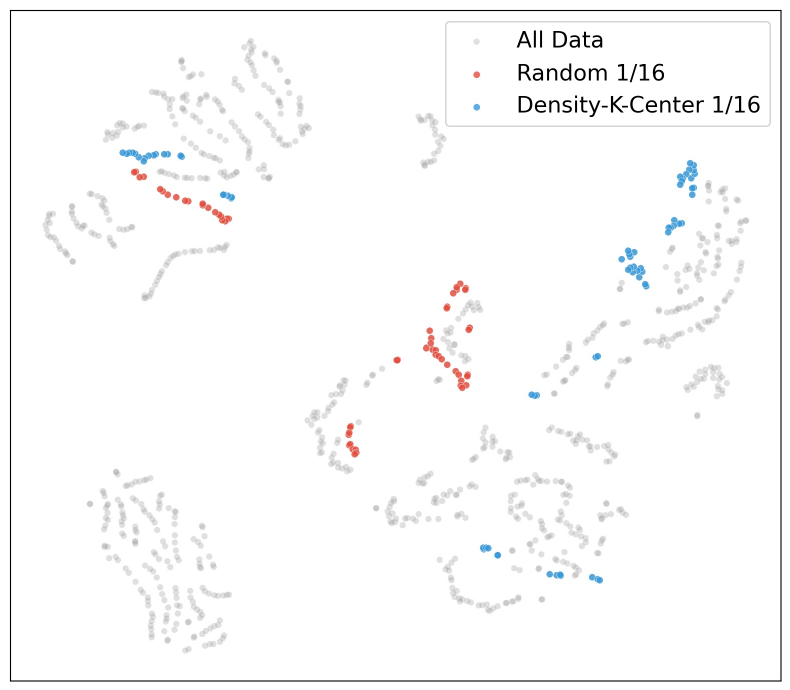}
  }
  \subfloat[Labeled ratio: 1/10]{
    \includegraphics[width=0.25\linewidth]{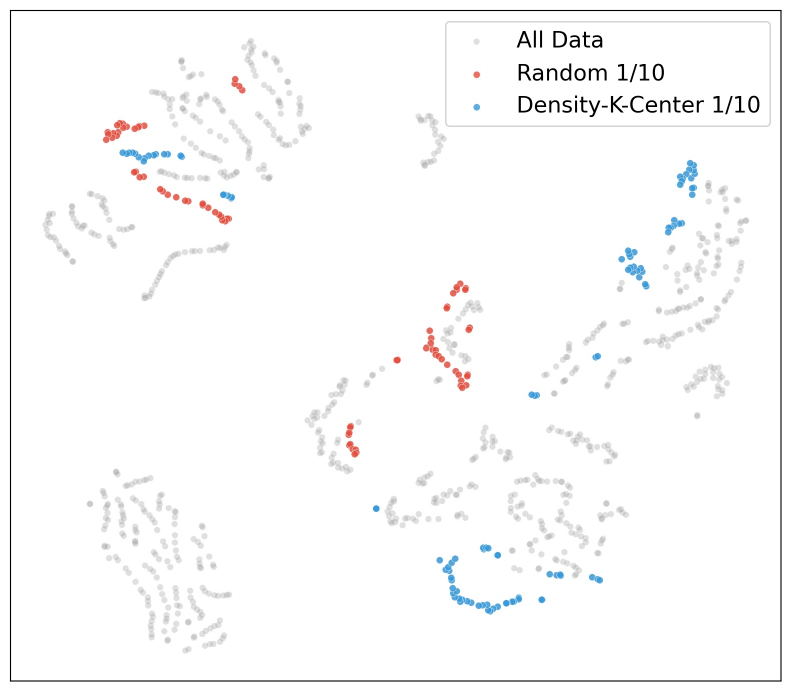}
  }
  \subfloat[Labeled ratio: 1/5]{
    \includegraphics[width=0.25\linewidth]{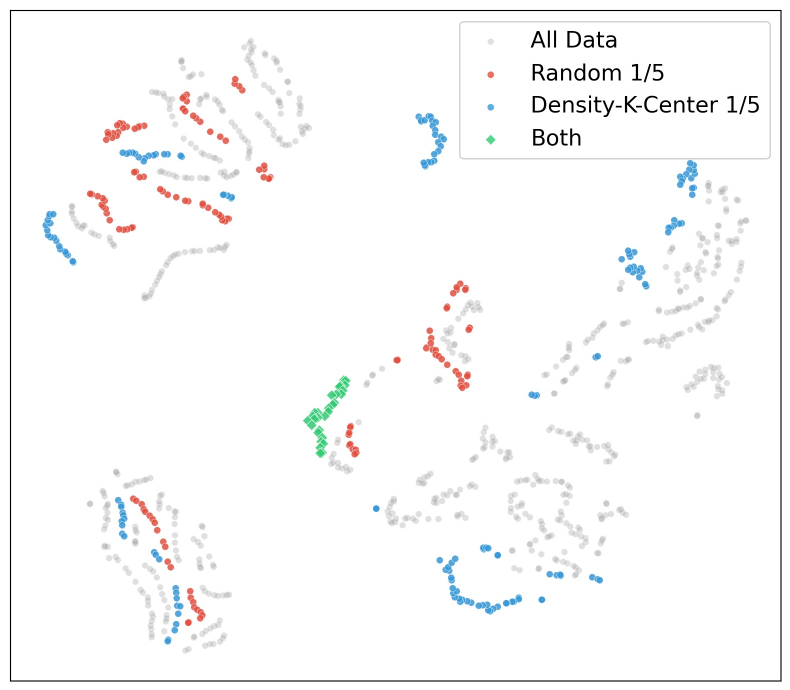}
  }
  \subfloat[Labeled ratio: 1/2]{
    \includegraphics[width=0.25\linewidth]{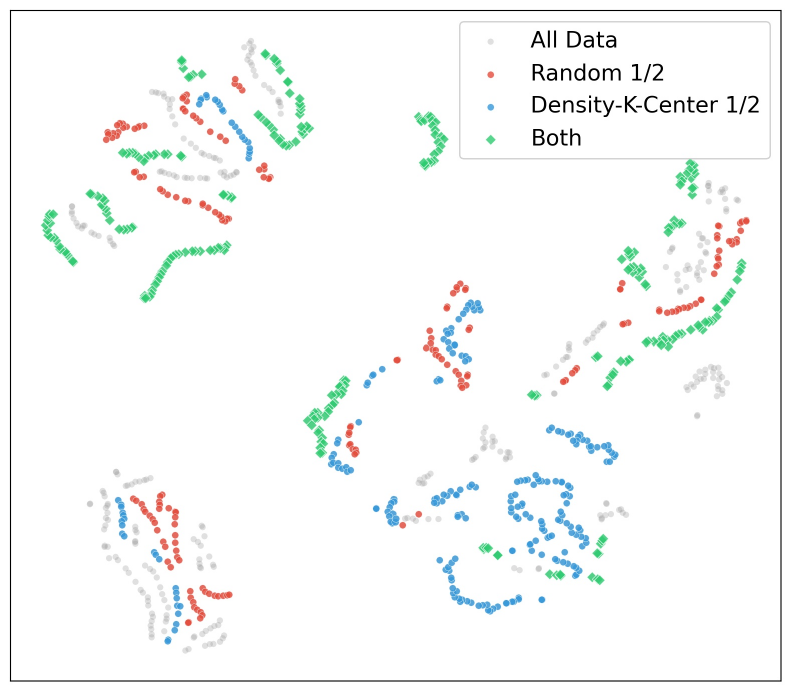}
  }}
  \vspace{-0.5em}
  \caption{t-SNE visualization of the PROMISE feature space. Background (gray) represents unlabeled data. Our Sample Selection (blue) covers broader feature-space clusters, while random sampling (red) shows localized clustering (representation bias).}
  \label{fig:tsne_promise12}
\end{figure}
\vspace{-2em}

\cref{fig:tsne_promise12} illustrates this geometric advantage on PROMISE. Standard random sampling suffers from representation bias, clustering locally. Conversely, our Density-K-Center strategy scatters anchors across broader regions of the data manifold, improving structural coverage even at a 1/16 ratio to reduce biased supervision.
{\color{rebuttalcolor}
To quantify this effect, we report feature-space coverage and labeled/unlabeled alignment diagnostics in \cref{tab:coverage_alignment}. The 95\% radius measures the distance from most unlabeled samples to their nearest anchor; cluster coverage/entropy measure the breadth and evenness of feature clusters; and L/U MMD$^2$ measures the labeled/unlabeled feature discrepancy after training. Across datasets, DKC improves anchor coverage over random sampling, and the full framework reduces feature discrepancy compared with UniV2.
}
\vspace{-1em}
\begin{table}[htbp]
\color{rebuttalcolor}
\centering
\caption{Feature-space coverage and alignment diagnostics. Coverage metrics compare Random avg.$\to$DKC; L/U MMD$^2$ compares UniV2$\to$Ours.}
\label{tab:coverage_alignment}
\setlength{\tabcolsep}{4pt}
\resizebox{\textwidth}{!}{
\begin{tabular}{llcccc}
\toprule
Dataset & Ratio & 95\% Rad.$\downarrow$ & Cluster Cov.$\uparrow$ & Cluster Ent.$\uparrow$ & L/U MMD$^2\downarrow$ \\
\midrule
BUSI & 1/8 & $0.0414 \!\to\! \textbf{0.0362}$ & $0.747 \!\to\! \textbf{0.860}$ & $0.885 \!\to\! \textbf{0.930}$ & $0.0100 \!\to\! \textbf{0.0085}$ \\
CAMUS & 1/8 & $0.0029 \!\to\! \textbf{0.0023}$ & $0.967 \!\to\! \textbf{1.000}$ & $0.941 \!\to\! \textbf{0.970}$ & $0.0058 \!\to\! \textbf{0.0052}$ \\
PMTCXR & 1/16 & $0.0161 \!\to\! \textbf{0.0150}$ & $0.880 \!\to\! \textbf{0.980}$ & $0.930 \!\to\! \textbf{0.952}$ & $0.0388 \!\to\! \textbf{0.0233}$ \\
PROMISE & 1/10 & $0.0681 \!\to\! \textbf{0.0423}$ & $0.147 \!\to\! \textbf{0.300}$ & $0.445 \!\to\! \textbf{0.635}$ & $0.0836 \!\to\! \textbf{0.0444}$ \\
\bottomrule
\end{tabular}
}
\end{table}
\vspace{-2em}

\cref{tab:ablation_selection} quantitatively validates these feature extraction choices. For encoding, utilizing Patch tokens from four intermediate layers consistently outperforms (or performs highly competitively with) the Unlocalized CLS token across all data ratios. This multi-level embedding better preserves the fine-grained spatial semantics crucial for boundary refinement. Furthermore, compared to sampling baselines like K-Means\cite{kmeans}, Facility Location\cite{facility}, and K-Center\cite{k-center}, DKC shows robust scalability. Since all three ratios ($\leq$1/4) represent low-label regimes, the final Patch-Four-DKC configuration is selected by its best average IoU/DSC/ASD/HD performance across them, forming our default pipeline.
% We regard all ratios up to 1/4 as low-label regimes and choose the default setting based on its overall behavior across the three ratios.

\vspace{-1em}
\begin{table}[htbp] % 如果需要固定位置，可以改为 [H]
\centering
\caption{Ablation study on offline sample selection strategies. We compare different feature encoding methods (CLS, Patch, or both C+P), extracted layers (Last vs. Four intermediate layers), and sampling algorithms.
\vspace{-0.2em} 
\textbf{DKC}: \textbf{Density-K-Center}.}
\label{tab:ablation_selection}
\renewcommand{\arraystretch}{1.1} 
\setlength{\tabcolsep}{8pt} 
\resizebox{\textwidth}{!}{
\begin{tabular}{l c l cc cc cc}
\toprule
\multicolumn{3}{c}{{Selection Settings}} & \multicolumn{2}{c}{{1/16 Ratio}} & \multicolumn{2}{c}{{1/8 Ratio}} & \multicolumn{2}{c}{{1/4 Ratio}} \\
\cmidrule(lr){1-3} \cmidrule(lr){4-5} \cmidrule(lr){6-7} \cmidrule(lr){8-9}
Encoding & Layers & Sampling & IoU/DSC$\uparrow$ & ASD/HD$\downarrow$ & IoU/DSC$\uparrow$ & ASD/HD$\downarrow$ & IoU/DSC$\uparrow$ & ASD/HD$\downarrow$ \\
\midrule
% --- Encoding Ablation ---
CLS       & Four & K-Center & \underline{72.0}/\textbf{79.0} & \textbf{26.9}/\textbf{37.4} & 72.0/78.5 & 30.6/39.2 & 75.0/81.0 & 26.5/34.5 \\
Patch     & Four & K-Center & \textbf{72.2}/\underline{78.7} & \underline{29.6}/\underline{39.3} & \underline{73.3}/\underline{79.7} & \textbf{29.0}/\underline{38.9} & \underline{75.4}/\underline{81.3} & \underline{24.6}/\underline{33.5} \\
C+P       & Four & K-Center & 70.8/77.3 & 34.9/45.1 & 71.4/77.7 & 31.3/41.3 & 74.9/80.9 & 26.3/35.5 \\
\midrule
% --- Layer Ablation ---
Patch     & Last & K-Center & 71.8/78.1 & 33.3/43.8 & 73.1/79.2 & 30.4/39.0 & 74.3/80.2 & 26.8/35.4 \\
\midrule
% --- Sampling Algorithm Comparison ---
Patch     & Four & K-Means  & 69.3/75.8 & 41.4/51.0 & 70.3/77.0 & 36.1/46.7 & 71.6/77.4 & 36.0/45.2 \\
Patch     & Four & Facility & 68.6/75.0 & 38.9/49.6 & 71.5/77.7 & 36.0/44.8 & 71.8/77.8 & 31.8/40.9 \\
\rowcolor{gray!10} % 突出显示最终采用的方法
Patch     & Four & DKC      & 71.5/78.1 & 30.6/40.9 & \textbf{74.1}/\textbf{79.8} & \underline{30.1}/\textbf{38.8} & \textbf{76.4}/\textbf{82.5} & \textbf{17.9}/\textbf{28.3} \\
\bottomrule
\end{tabular}
}
\end{table}
\vspace{-2em}

\subsection{Plug-and-Play Capability of Offline Sample Selection}

We evaluate the plug-and-play capability of our Distribution-Aware Sample Selection (SS) by integrating it into four diverse SSMIS methods (ABD, RCP, AdaMix, SynFoc) across varying backbones. As shown in \cref{tab:plug_and_play}, replacing random sampling with our explicitly selected anchors consistently improves segmentation overlap across all baselines on the BUSI dataset. Improvements are particularly noticeable in the scarce 1/16 regime, yielding DSC gains for AdaMix (+6.5\%) and SynFoc (+5.5\%), alongside reductions in boundary errors (\eg, an 18.9-pixel HD95 drop for AdaMix). This cross-architecture robustness suggests that capturing the global feature manifold can serve as a versatile performance multiplier for downstream semi-supervised learning.

{\color{rebuttalcolor}
\noindent\textbf{Stage-1 practicality.}
DKC ranks samples once using frozen VFM features and requires no query or retraining rounds. This is practical in medical scenarios, where large unlabeled archives are available and can be screened offline before annotation to build a representative labeled set. On BUSI/PMTCXR, feature extraction takes 2.4/8.9 minutes, DKC selection takes 0.5/0.6 seconds, and the Stage-1 cost is 2.9\%/2.1\% of one SSMIS training run, with no inference overhead.

\noindent\textbf{Active learning comparison.}
Unlike online active learning, DKC is a one-shot offline selector requiring no warm-up or retraining under identical annotation budgets. On 1/16 BUSI/PMTCXR, it achieves 78.1/48.7 \% DSC in just 0.5/0.6s, surpassing both BADGE~\cite{badge} (78.0/47.8) and BALD~\cite{bald} (77.4/46.3).
}

\vspace{-1em}
\begin{table}[htbp]
\centering
\caption{Plug-and-play performance of our Sample Selection (SS). The maximum improvement in each column is highlighted in \textbf{bold}, and the second-best is \underline{underlined}. Positive impacts are in \textcolor{green!60!black}{green}, while negative impacts are in \textcolor{red}{red}.}
\vspace{-0.2em}
\label{tab:plug_and_play}
\setlength{\tabcolsep}{4pt}
\resizebox{\textwidth}{!}{
\begin{tabular}{l c c cc cc cc}
\toprule
\multirow{3}{*}{Methods} & \multirow{3}{*}{Venue} & \multirow{3}{*}{Encoder} & \multicolumn{6}{c}{Ratio and the number of labeled images} \\
\cmidrule(lr){4-9}
& & & \multicolumn{2}{c}{1/16 (39)} & \multicolumn{2}{c}{1/8 (78)} & \multicolumn{2}{c}{1/4 (156)} \\
\cmidrule(lr){4-5} \cmidrule(lr){6-7} \cmidrule(lr){8-9}
& & & IoU/DSC$\uparrow$ & ASD/HD$\downarrow$ & IoU/DSC$\uparrow$ & ASD/HD$\downarrow$ & IoU/DSC$\uparrow$ & ASD/HD$\downarrow$ \\
\midrule
% ==================== ABD ====================
ABD \cite{ABD} & \multirow{3}{*}{CVPR'24} & \multirow{3}{*}{SwinUNet} & 63.6/71.0 & 42.0/57.6 & 66.3/73.3 & 39.5/53.5 & 69.9/76.4 & 33.1/45.0 \\
+ our SS & & & 65.8/73.3 & 39.9/53.5 & 68.3/75.1 & 44.0/54.7 & 71.4/78.1 & 34.1/44.7 \\ 
$\Delta$ & & & \textcolor{green!60!black}{$\uparrow$2.2}/\textcolor{green!60!black}{$\uparrow$2.3} & \underline{\textcolor{green!60!black}{$\downarrow$2.1}}/\underline{\textcolor{green!60!black}{$\downarrow$4.1}} & \underline{\textcolor{green!60!black}{$\uparrow$2.0}}/\textcolor{green!60!black}{$\uparrow$1.8} & \textcolor{red}{$\uparrow$4.5}/\textcolor{red}{$\uparrow$1.2} & \underline{\textcolor{green!60!black}{$\uparrow$1.5}}/\underline{\textcolor{green!60!black}{$\uparrow$1.7}} & \textcolor{red}{$\uparrow$1.0}/\textcolor{green!60!black}{$\downarrow$0.3} \\
\midrule
% ==================== RCP ====================
RCP \cite{RCP} & \multirow{3}{*}{TIP'25} & \multirow{3}{*}{UNet} & 64.3/71.2 & 44.2/57.8 & 64.9/71.9 & 42.4/55.4 & 65.1/71.8 & 38.6/53.5 \\
+ our SS & & & 65.5/72.5 & 42.9/55.2 & 67.9/75.0 & 39.9/50.7 & 68.6/75.2 & 41.0/51.7 \\ 
$\Delta$ & & & \textcolor{green!60!black}{$\uparrow$1.2}/\textcolor{green!60!black}{$\uparrow$1.3} & \textcolor{green!60!black}{$\downarrow$1.3}/\textcolor{green!60!black}{$\downarrow$2.6} & \textbf{\textcolor{green!60!black}{$\uparrow$3.0}}/\textbf{\textcolor{green!60!black}{$\uparrow$3.1}} & \underline{\textcolor{green!60!black}{$\downarrow$2.5}}/\underline{\textcolor{green!60!black}{$\downarrow$4.7}} & \textbf{\textcolor{green!60!black}{$\uparrow$3.5}}/\textbf{\textcolor{green!60!black}{$\uparrow$3.4}} & \textcolor{red}{$\uparrow$2.4}/\textcolor{green!60!black}{$\downarrow$1.8} \\
\midrule
% ==================== AdaMix ====================
AdaMix \cite{AdaMix} & \multirow{3}{*}{MedIA'25} & \multirow{3}{*}{UNet} & 51.6/58.6 & 65.2/79.3 & 61.0/67.9 & 49.7/60.9 & 63.2/70.0 & 49.5/58.9 \\
+ our SS & & & 58.5/65.1 & 45.6/60.4 & 62.6/68.9 & 50.8/62.6 & 64.6/71.5 & 42.7/52.6 \\ 
$\Delta$ & & & \textbf{\textcolor{green!60!black}{$\uparrow$6.9}}/\textbf{\textcolor{green!60!black}{$\uparrow$6.5}} & \textbf{\textcolor{green!60!black}{$\downarrow$19.6}}/\textbf{\textcolor{green!60!black}{$\downarrow$18.9}} & \textcolor{green!60!black}{$\uparrow$1.6}/\textcolor{green!60!black}{$\uparrow$1.0} & \textcolor{red}{$\uparrow$1.1}/\textcolor{red}{$\uparrow$1.7} & \textcolor{green!60!black}{$\uparrow$1.4}/\textcolor{green!60!black}{$\uparrow$1.5} & \underline{\textcolor{green!60!black}{$\downarrow$6.8}}/\underline{\textcolor{green!60!black}{$\downarrow$6.3}} \\
\midrule
% ==================== SynFoc ====================
SynFoc \cite{synfoc} & \multirow{3}{*}{CVPR'25} & \multirow{3}{*}{MedSAM} & 64.8/71.3 & 39.2/54.0 & 69.0/74.9 & 35.4/47.0 & 71.8/78.1 & 27.3/37.3 \\
+ our SS & & & 70.5/76.8 & 40.9/50.3 & 70.2/76.8 & 27.5/39.7 & 72.8/79.4 & 20.4/30.8 \\
$\Delta$ & & & \underline{\textcolor{green!60!black}{$\uparrow$5.7}}/\underline{\textcolor{green!60!black}{$\uparrow$5.5}} & \textcolor{red}{$\uparrow$1.7}/\textcolor{green!60!black}{$\downarrow$3.7} & \textcolor{green!60!black}{$\uparrow$1.2}/\underline{\textcolor{green!60!black}{$\uparrow$1.9}} & \textbf{\textcolor{green!60!black}{$\downarrow$7.9}}/\textbf{\textcolor{green!60!black}{$\downarrow$7.3}} & \textcolor{green!60!black}{$\uparrow$1.0}/\textcolor{green!60!black}{$\uparrow$1.3} & \textbf{\textcolor{green!60!black}{$\downarrow$6.9}}/\textbf{\textcolor{green!60!black}{$\downarrow$6.5}} \\
\bottomrule
\end{tabular}
}
\end{table}
\vspace{-1em}

\vspace{-1em}

\subsection{Comparison with Alternative CP Strategies}
\label{sec:alternative_cp}

% \cref{tab:alternative_cp} compares Copy-Paste strategies under identical UniV2\cite{unimatchv2} baseline and splits. Traditional methods often introduce boundary artifacts, causing severe degradation in low-data regimes (\eg, AdaMix's HD95 collapse to 51.07 at 1/16). In contrast, MCP achieves the highest DSC (78.29\%) at 1/16 and keeps HD95 below 40 pixels. At 1/4, MCP maintains the best DSC (80.52\%). While BCP yields a marginally lower ASD at 1/4, MCP provides stronger robustness against severe boundary failures (best HD95). This suggests that our Memory Bank can replace blind spatial mixing with more reliable semantic anchors, preserving anatomical integrity.

\cref{tab:alternative_cp} compares Copy-Paste strategies under identical UniV2\cite{unimatchv2} baseline and splits. Traditional methods introduce boundary artifacts, causing severe degradation in low-data regimes (\eg, AdaMix's HD95 collapse to 51.07 at 1/16). Conversely, MCP achieves the highest DSC (78.29\%) at 1/16 and keeps HD95 below 40 pixels. At 1/4, MCP maintains the best DSC (80.52\%). While BCP yields a marginally lower ASD at 1/4, MCP enhances robustness against severe boundary failures (best HD95). This suggests Memory Bank replaces blind spatial mixing with more reliable semantic anchors, preserving anatomical integrity.

% \vspace{-1.5em}
\begin{table}[htbp]
\centering
\caption{Comparison of copy-paste strategies. All methods use the same UniV2 baseline and {DINOv2-S} encoder to ensure a fair evaluation of the mixing strategy itself.}
\vspace{-0.25em}
\label{tab:alternative_cp}
\setlength{\tabcolsep}{3pt} % 稍微收紧列间距以容纳新增的Encoder列
\resizebox{\textwidth}{!}{
\begin{tabular}{l c c cccc cccc}
\toprule
\multirow{2}{*}{Strategies} & \multirow{2}{*}{Venue} & \multirow{2}{*}{Encoder} & \multicolumn{4}{c}{1/16 Ratio} & \multicolumn{4}{c}{1/4 Ratio} \\
\cmidrule(lr){4-7} \cmidrule(lr){8-11}
& & & IoU(\%)$\uparrow$ & DSC(\%)$\uparrow$ & ASD$\downarrow$ & HD95$\downarrow$ & IoU(\%)$\uparrow$ & DSC(\%)$\uparrow$ & ASD$\downarrow$ & HD95$\downarrow$ \\
\midrule
BCP \cite{BCP}       & CVPR'23 & \multirow{5}{*}{DINOv2-S} & 68.62 & 75.51 & \underline{27.03} & \underline{40.36} & \underline{73.97} & \underline{80.20} & \textbf{23.95} & \underline{33.85} \\
TP-RAM \cite{midss}  & CVPR'24 &  & 69.43 & 76.34 & 28.72 & 40.98 & 73.67 & 79.58 & 28.68 & 37.14 \\
ABD \cite{ABD}       & CVPR'24 &  & \underline{70.51} & \underline{77.05} & 30.44 & 41.92 & 72.47 & 78.54 & 31.55 & 40.16 \\
RCP \cite{RCP}       & TIP'25  &  & 69.41 & 75.92 & 30.14 & 43.31 & 73.40 & 79.29 & 27.10 & 35.85 \\
AdaMix \cite{AdaMix} & MedIA'25&   & 67.92 & 74.44 & 39.72 & 51.07 & 73.08 & 78.69 & 27.44 & 37.05 \\
\midrule
\textbf{MCP (Ours)}  & -       & DINOv2-S & \textbf{71.76} & \textbf{78.29} & \textbf{26.14} & \textbf{37.84} & \textbf{74.76} & \textbf{80.52} & \underline{25.07} & \textbf{33.49} \\
\bottomrule
\end{tabular}
}
\end{table}
% \vspace{-1.5em}

\subsection{Hyperparameter Analysis of MCP}
\begin{figure}[htbp]
  \centering
  % \vspace{-1em}
  \subfloat[$C_{bank}$]{
    \includegraphics[height=0.225\linewidth]{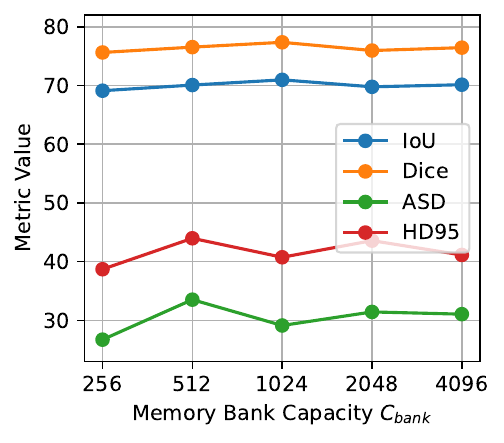}
  }
  \subfloat[$P_{bank}$]{
    \includegraphics[height=0.225\linewidth]{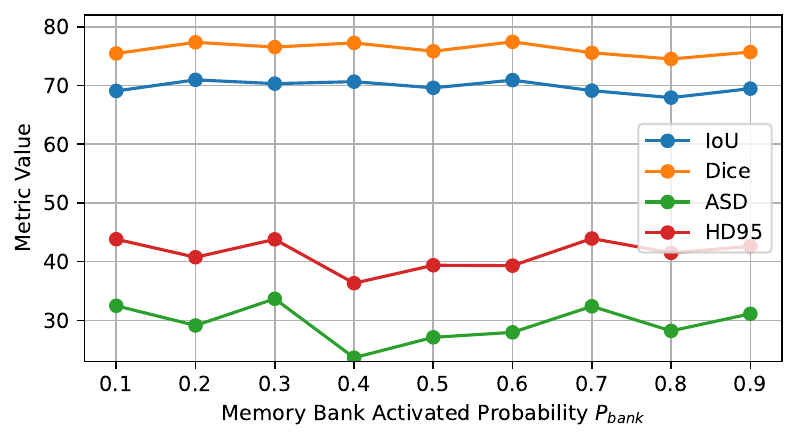}
  }
  \subfloat[$R_{prog}$]{
    \includegraphics[height=0.225\linewidth]{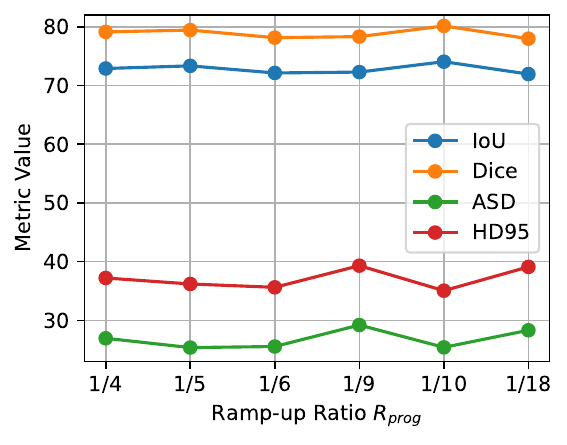}
  }
  % \vspace{-0.5em}
  \caption{Performance of MCP module under different hyperparameters on BUSI dataset. (a) Bank Capacity $C_{bank}$. (b) Activation Probability $P_{bank}$. (c) Ramp-up Ratio $R_{prog}$.}
  \label{fig:hyper_param}
\end{figure}

\cref{fig:hyper_param} details MCP's hyperparameter sensitivity on BUSI. %($C_{bank}$, $P_{bank}$: 1/16, $R_{prog}$: 1/8).

\textbf{Bank Capacity ($C_{bank}$) at 1/16 ratio.}
A small capacity ($256$) restricts diversity, while a large bank ($\ge 2048$) retains outdated, low-quality patches. Setting $C_{bank}=1024$ balances diversity and reliability (77.35\% DSC).

\textbf{Retrieval Probability ($P_{bank}$) at 1/16 ratio.}
A low probability ($0.1$) underutilizes memory, whereas high values ($\ge 0.8$) force over-reliance on historical features, dropping DSC to 74.51\%. $P_{bank}=0.2$ provides the good trade-off.

\textbf{Progressive Ratio ($R_{prog}$) at 1/8 ratio.} 
An aggressive schedule ($1/18$) prematurely introduces complex augmentations, risking overfitting to early noise. Conversely, a prolonged schedule ($1/4$) delays regularization. The $1/10$ ratio safely navigates early stages, peaking at 80.14\% DSC and 35.06 HD95.

\vspace{-1em}
\subsection{Qualitative Analysis}
\begin{figure}[H]
  \centering
  
  % --- 第一行：BUSI ---
  \begin{minipage}[c]{0.03\textwidth}
    \centering
    % \raisebox{0pt}[0pt][0pt] 用于彻底消除文字的垂直占位，防止撑爆行距
    \raisebox{6.5pt}[0pt][0pt]{\rotatebox{90}{\scriptsize \textbf{BUSI}}}
  \end{minipage}%
  \hfill
  \subfloat{\begin{minipage}[c]{0.47\textwidth}
    \centering
    \includegraphics[width=\linewidth]{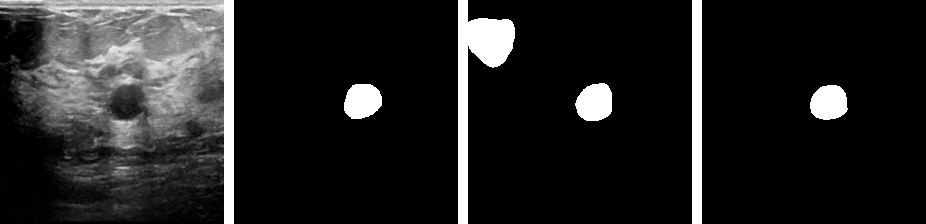}
  \end{minipage}}
  \hfill
  \subfloat{\begin{minipage}[c]{0.47\textwidth}
    \centering
    \includegraphics[width=\linewidth]{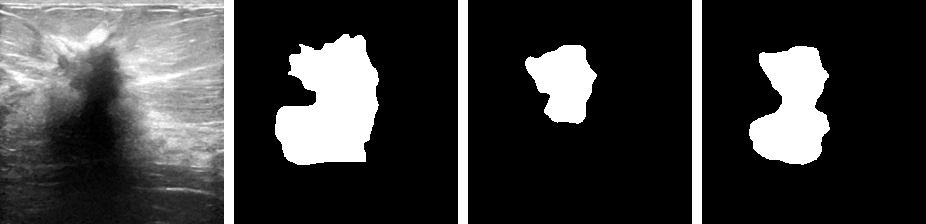}
  \end{minipage}}\\
  \vspace{0.5mm} % 稍微给一点点自然间距
  
  % --- 第二行：PMTCXR ---
  \begin{minipage}[c]{0.03\textwidth}
    \centering
    \raisebox{-1pt}[0pt][0pt]{\rotatebox{90}{\scriptsize \textbf{PMTCX.}}}
  \end{minipage}%
  \hfill
  \subfloat{\begin{minipage}[c]{0.47\textwidth}
    \centering
    \includegraphics[width=\linewidth]{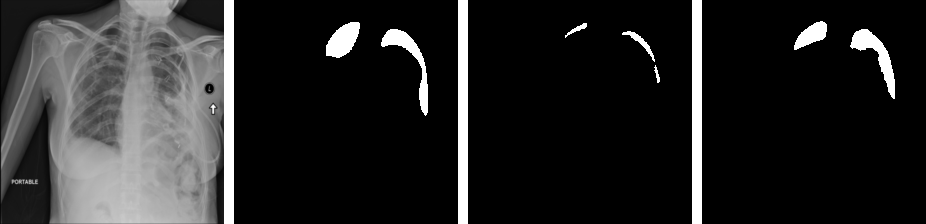}
  \end{minipage}}
  \hfill
  \subfloat{\begin{minipage}[c]{0.47\textwidth}
    \centering
    \includegraphics[width=\linewidth]{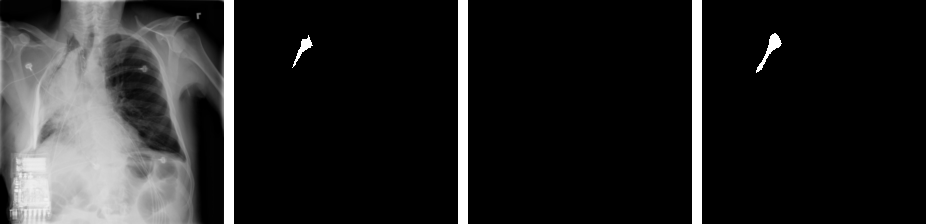}
  \end{minipage}}\\
  \vspace{0.5mm}
  
  % --- 第三行：ISIC ---
  \begin{minipage}[c]{0.03\textwidth}
    \centering
    \raisebox{8pt}[0pt][0pt]{\rotatebox{90}{\scriptsize \textbf{ISIC}}}
  \end{minipage}%
  \hfill
  \subfloat{\begin{minipage}[c]{0.47\textwidth}
    \centering
    \includegraphics[width=\linewidth]{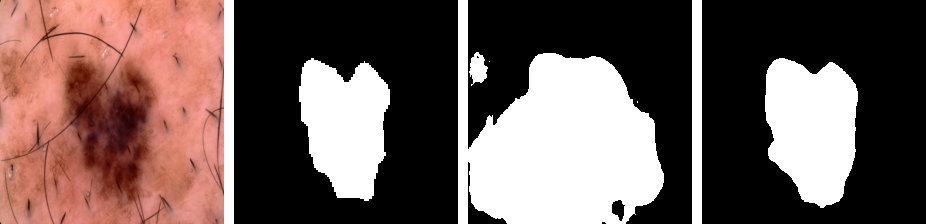}
  \end{minipage}}
  \hfill
  \subfloat{\begin{minipage}[c]{0.47\textwidth}
    \centering
    \includegraphics[width=\linewidth]{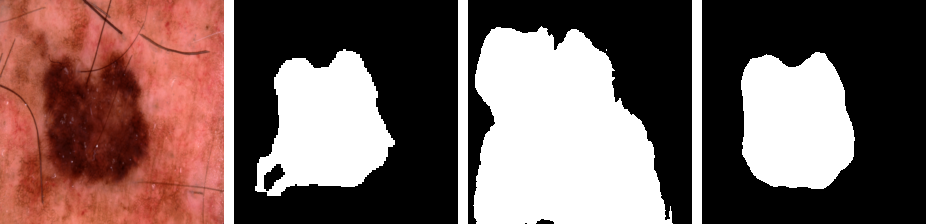}
  \end{minipage}}\\
  \vspace{0.5mm}
  
  % --- 第四行：CAMUS ---
  \begin{minipage}[c]{0.03\textwidth}
    \centering
    \raisebox{0pt}[0pt][0pt]{\rotatebox{90}{\scriptsize \textbf{CAMUS}}}
  \end{minipage}%
  \hfill
  \subfloat{\begin{minipage}[c]{0.47\textwidth}
    \centering
    \includegraphics[width=\linewidth]{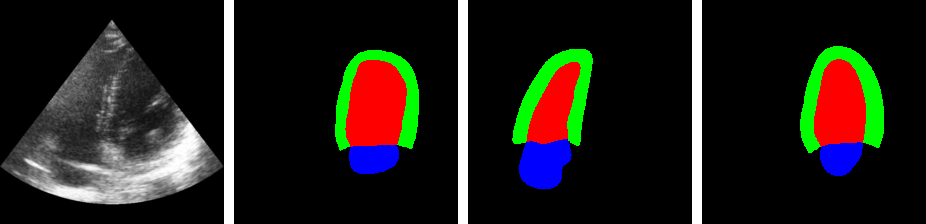}
  \end{minipage}}
  \hfill
  \subfloat{\begin{minipage}[c]{0.47\textwidth}
    \centering
    \includegraphics[width=\linewidth]{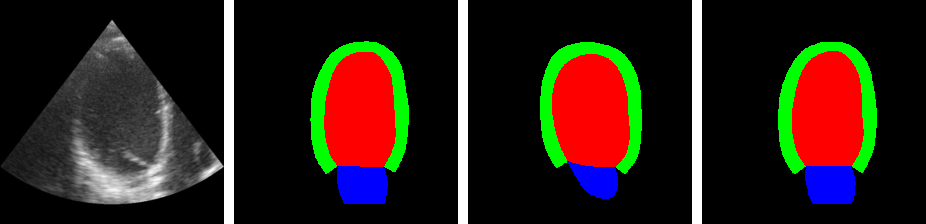}
  \end{minipage}}\\
  \vspace{0.5mm}

  % --- 第五行：PROMISE ---
  \begin{minipage}[c]{0.03\textwidth}
    \centering
    \raisebox{1.5pt}[0pt][0pt]{\rotatebox{90}{\scriptsize \textbf{PROMI.}}}
  \end{minipage}%
  \hfill
  \subfloat{\begin{minipage}[c]{0.47\textwidth}
    \centering
    \includegraphics[width=\linewidth]{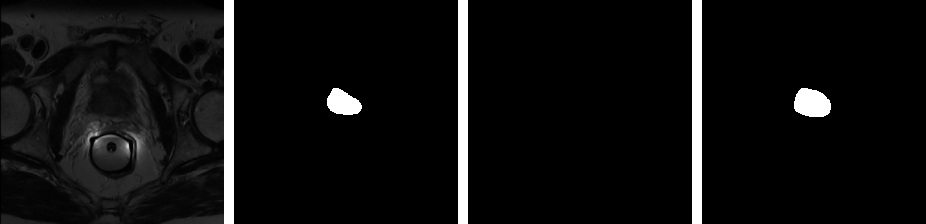}
  \end{minipage}}
  \hfill
  \subfloat{\begin{minipage}[c]{0.47\textwidth}
    \centering
    \includegraphics[width=\linewidth]{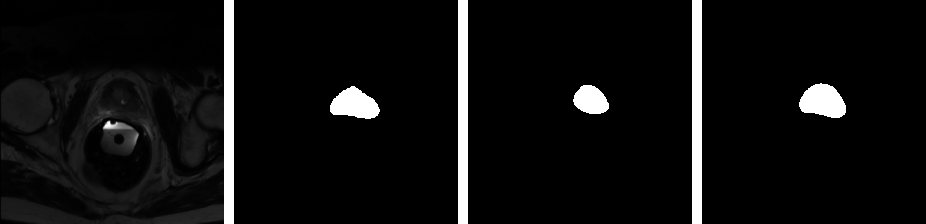}
  \end{minipage}}\\
  \vspace{0.5mm} 

  % --- 第六行：ACDC ---
  \begin{minipage}[c]{0.03\textwidth}
    \centering
    \raisebox{4pt}[0pt][0pt]{\rotatebox{90}{\scriptsize \textbf{ACDC}}}
  \end{minipage}%
  \hfill
  \subfloat{\begin{minipage}[c]{0.47\textwidth}
    \centering
    \includegraphics[width=\linewidth]{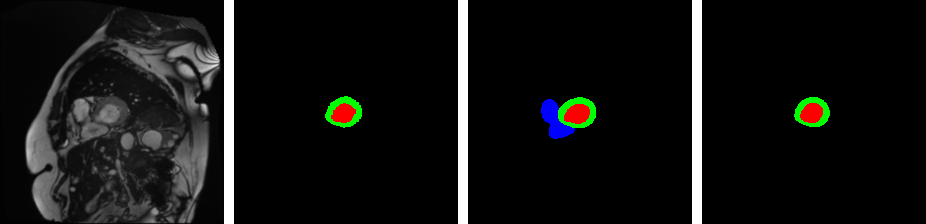}
  \end{minipage}}
  \hfill
  \subfloat{\begin{minipage}[c]{0.47\textwidth}
    \centering
    \includegraphics[width=\linewidth]{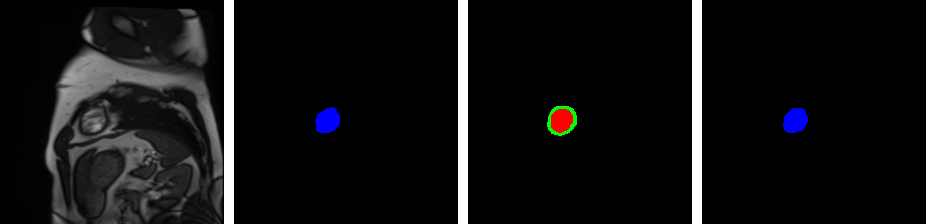}
  \end{minipage}}\\
  \vspace{1mm} 
  
  % --- 最下方的文字标签行 ---
  \begin{minipage}[c]{0.03\textwidth}
    \hfill
  \end{minipage}%
  \hfill
  \subfloat{\begin{minipage}[c]{0.47\textwidth}
    \centering
    \small 
    \makebox[0.25\linewidth]{Image}%
    \makebox[0.25\linewidth]{GT}%
    \makebox[0.25\linewidth]{UniV2}%
    \makebox[0.25\linewidth]{Ours}
  \end{minipage}}
  \hfill
  \subfloat{\begin{minipage}[c]{0.47\textwidth}
    \centering
    \small
    \makebox[0.25\linewidth]{Image}%
    \makebox[0.25\linewidth]{GT}%
    \makebox[0.25\linewidth]{UniV2}%
    \makebox[0.25\linewidth]{Ours}
  \end{minipage}}
  
  % \vspace{-0.5em}
  \caption{Qualitative results on BUSI, PMTCXR, ISIC, CAMUS, PROMISE and ACDC (top to bottom). Columns display the Image, Ground Truth (GT), UniV2, and Ours.}
  \label{fig:qualitative_results}
\end{figure}

\cref{fig:qualitative_results} shows segmentation results across six medical datasets. Compared to the baseline UniV2, our method demonstrates improved precision. In challenging cases with small structures (PMTCXR, PROMISE, ACDC), UniV2 often under-segments or misses target regions, whereas our approach captures them more reliably. For complex lesions (BUSI, ISIC, CAMUS), we produce smoother, more accurate boundaries aligned with the ground truth.

% \vspace{-1em}
% \vspace{-3em}

% \subsection{Discussion}

\section{Conclusion}
% \vspace{-0.5em}
We propose a data-efficient Semi-Supervised Medical Image Segmentation framework tackling representation bias and pseudo-label noise in extreme low-data regimes. First, Distribution-Aware Sample Selection captures the global anatomical manifold, preventing local minima collapse. Second, the Memory-guided Copy-Paste (MCP) module suppresses noisy pseudo-labels by retrieving historical patches via a KL-guided Semantic Memory Bank. Coupled with progressive training, MCP robustly refines boundaries and preserves structural integrity. 
Experiments across six datasets show competitive segmentation performance, even at 1/16 labeled ratios. Despite training overhead, inference remains zero-cost. While currently processing 3D data slice-by-slice, future work will extend MCP to direct 3D volume processing and adaptive memory management.

% \clearpage\mbox{}Page \thepage\ of the manuscript.
% \clearpage\mbox{}Page \thepage\ of the manuscript.
% \clearpage\mbox{}Page \thepage\ of the manuscript.
% \clearpage\mbox{}Page \thepage\ of the manuscript.
% \clearpage\mbox{}Page \thepage\ of the manuscript. This is the last page.
% Now we have reached the maximum length of an ECCV \ECCVyear{} submission (excluding references and acknowledgements).
% References should start immediately after the main text, but can continue past p.\ 14 if needed. 

\section*{Acknowledgements}
% \noindent\textbf{Acknowledgements.}
% Please insert your acknowledgments here.
This work is supported by the National Natural Science Foundation of China (Grants No. 62473253, U22A20100).
\bibliographystyle{splncs04}
\bibliography{main}
\end{document}